\documentclass{article}

\usepackage{microtype}
\usepackage{graphicx}
\usepackage{booktabs} 

\usepackage{amsmath}
\usepackage{amssymb}
\usepackage{xcolor}
\usepackage{algorithm}
\usepackage{algorithmic}
\usepackage{pgfplots}
\usepackage{multirow}
\usepackage{array}
\usepackage{subcaption}
\usepackage{diagbox, eqparbox, hhline}
\usepackage[normalem]{ulem}
\usepackage{hyperref}

\graphicspath{{figures/}}

\newcommand{\softa}{{\ttfamily BoXHED} }
\newcommand{\softb}{{\ttfamily BoXHED}}
\newcommand{\xgb}{{\ttfamily XGBoost} }

\newcommand{\flexsurv}{{\ttfamily flexsurv} }
\newcommand{\flexsurvb}{{\ttfamily flexsurv}}
\newcommand{\kernel}{{\ttfamily kernel} }
\newcommand{\kernelb}{{\ttfamily kernel}}
\newcommand{\blackboost}{{\ttfamily blackboost} }

\newcommand{\eqindent}{\hspace{2em}}
\newcommand{\tabincell}[2]{\begin{tabular}{@{}#1@{}}#2\end{tabular}}

\usepackage[accepted]{icml2020}

\icmltitlerunning{\softb: Boosted eXact Hazard Estimator with Dynamic covariates}

\begin{document}

\twocolumn[
\icmltitle{\softb: Boosted eXact Hazard Estimator with Dynamic covariates}



\icmlsetsymbol{equal}{*}

\begin{icmlauthorlist}
\icmlauthor{Xiaochen Wang}{yale}
\icmlauthor{Arash Pakbin}{tamu}
\icmlauthor{Bobak J. Mortazavi}{tamu}
\icmlauthor{Hongyu Zhao}{yale}
\icmlauthor{Donald K.K. Lee}{emory}
\end{icmlauthorlist}

\icmlaffiliation{yale}{Biostatistics Department, Yale University, New Haven, Connecticut, USA}
\icmlaffiliation{tamu}{Computer Science \& Engineering, Texas A\&M University, College Station, Texas, USA}
\icmlaffiliation{emory}{Goizueta Business School and Department of Biostatistics \& Bioinformatics, Emory University, Atlanta, Georgia, USA}

\icmlcorrespondingauthor{Donald K.K. Lee}{donald.lee@emory.edu}

\icmlkeywords{Survival Analysis, Gradient Boosting, Nonparametric Hazard Estimation, Functional Data}

\vskip 0.3in
]



\printAffiliationsAndNotice{}  

\begin{abstract}
The proliferation of medical monitoring devices makes it possible to track health vitals at high frequency, enabling the development of dynamic health risk scores that change with the underlying readings. Survival analysis, in particular hazard estimation, is well-suited to analyzing this stream of data to predict disease onset as a function of the time-varying vitals. This paper introduces the software package \softa (pronounced `box-head') for nonparametrically estimating hazard functions via gradient boosting. \softa 1.0 is a novel tree-based implementation of the generic estimator proposed in \citet{lee2017boosted}, which was designed for handling time-dependent covariates in a fully nonparametric manner. \softa is also the first publicly available software implementation for \citet{lee2017boosted}. Applying \softa to cardiovascular disease onset data from the Framingham Heart Study reveals novel interaction effects among known risk factors, potentially resolving an open question in clinical literature.
\end{abstract}

\section{Introduction}
\label{submission}

Driven by numerous applications in healthcare data analytics, interest in machine learning methods for survival data is rising steadily. These machine learning techniques often focus on classification problems (e.g. binary prediction of mortality in the next 72 hours), but we may be interested in not only the probability of an event occurring, but also in when that event might occur. For patients admitted to an Intensive Care Unit (ICU), for example, the high frequency nature of data collected by electronic health record systems allows us to prognosticate not only the chance of a patient dying, but also when that might happen, in order to provide timely critical care. Another example is the Framingham Heart Study, where longitudinal measurements allow us to ask not only if a patient will develop cardiovascular diseases (CVDs), but also when they are likely to develop them. In these settings, uncovering complex interactions of the dynamics of these risk factors could help quantify real-time risk and enable life-saving care. We explore an approach to this time-to-event analysis, called survival analysis in the statistics literature, to improve upon existing techniques by accounting for the time-varying nature of patient risk factors and potentially addressing new clinical findings.

Survival data describe the time to an event of interest (e.g. patient mortality), and a characteristic feature is the presence of censoring and/or truncation of the actual time-to-event $T$. This loss of information makes survival analysis fundamentally different from learning from continuous-valued data on the real line. For example, in lieu of the density $\mathbb{P}(T\in dt)$ and cumulative distribution $\mathbb{P}(T\le t)$, the hazard $\lambda(t)=\mathbb{P}(T\in dt|T>t)$ and survivor function $S(t)=\mathbb{P}(T>t)$ are more natural for the survival setting. Recent works for survival analysis centre on methods for estimating the conditional survivor function
\begin{equation}\label{eq:cond_surv}
S(t|X) =  \mathbb{P}(T>t|X) = \exp\left(-\int_0^t \lambda(u,X)du\right)
\end{equation}
given time-static covariates $X\in\mathbb{R}^p$ that are fixed at $t=0$.

In many real world settings, the covariates $X(t)\in\mathbb{R}^p$ can in fact change over time (e.g. patient temperature), and thanks to advances in technology, their readings can now be captured in real time. When $X(t)$ is time-varying there is no meaningful analogue to the survivor function \eqref{eq:cond_surv}, since it involves integrating the hazard $\lambda(u,X(u))$ along the unknown future trajectory of the covariates $\{X(u)\}_{u\in(0,t]}$. Thus in the time-dependent covariate setting, we are instead interested in estimating the hazard $\lambda(t,x)$ given $X(t)=x$, which is informally the probability of the event happening in the near future given it has not happened yet. Note that in the time-static covariate setting, we can recover $S(t|x)$ from $\lambda(t,x)$ via \eqref{eq:cond_surv}. Therefore, by focusing on hazard estimation we can unify the analyses of both settings.

However, nonparametric hazard estimation in the time-dependent covariate setting is challenging because the covariate trajectories of the observations are functional data points. This may explain the sparsity of the literature on this topic. A recent work by \citet{lee2017boosted} examined this problem. Specifically, they proposed a functional gradient boosting algorithm for estimating the hazard nonparametrically using arbitrary weak learners. Theoretical guarantees are provided for the estimator, and they also outlined a prototype tree-based implementation as proof-of-concept. The authors deferred the development of a more refined implementation for future research, and hence did not provide software.

In order to design a scalable implementation that is clinically and software deployable, and to extend the machine learning toolbox for survival analysis to the time-dependent covariate setting, we introduce the algorithm and software package \softa (pronounced `box-head'). This is the first publicly available implementation of the generic hazard estimator proposed in \citet{lee2017boosted}. The current version, \softa 1.0 (\href{www.github.com/BoXHED/BoXHED1.0}{www.github.com/BoXHED/BoXHED1.0}), is written in Python and uses regression trees as learners. We first describe the novel algorithmic aspects of our implementation vis-\`a-vis the tree-based prototype in \citet{lee2017boosted}. The key innovation of \softa is in developing a novel approach for growing trees that is more targeted at likelihood risk reduction, and as a result gives rise to a more direct variable importance measure. We evaluate the performance of \softa on simulation experiments, and also use it to analyze a cardiovascular disease dataset from the Framingham Heart Study. Of note, \softa discovered novel interaction effects among known risk factors in the Framingham data, potentially resolving an open question in the clinical literature. This further illustrates the utility of our tool for performing healthcare data analytics.

\section{Related Work}
The machine learning literature on survival analysis to date focus mainly on the time-static covariate setting \citep{ishwaran2008random, ranganath2016deep, bellot2018boosted, bellot2019boosting, lee2019temporal}. There is, however, growing awareness of the importance of time-dependent covariates. For example, neural networks have been developed for hazard modeling in a discrete-time setting \citep{jarrett2018match,ren2019}. This is equivalent to solving a series of binary classification problems, one at each point in an equally spaced time-grid. We note that there is also literature on neural network survival models that are not hazard-based, also for the discrete-time setting.

The extension to hazard modeling in a continuous-time setting has a longer history in the statistics literature. For time-static covariates, boosting has been applied to both parametric hazard models \citep{buhlmann2007} as well as the semiparametric Cox proportional hazards model \citep{ridgeway1999state,li2005boosting,binder2008allowing}. Where time-dependent covariates are concerned, kernel smoothing estimators \citep{neilsen,perez} have been proposed for low-dimensional covariate settings (along with theoretical guarantees), but otherwise the literature is rather sparse. 


\section{Problem Description}\label{loss_function}
Consider functional datapoints collected from $n$ units, with the $i$-th one represented by  
$$
(X_{i}(t)_{t\le \tilde T_i},\tilde T_{i},\Delta_{i}), 
$$
where $\tilde T_i$ is the minimum of the event time $T_i$ and censoring time $C_i$, $\Delta_i=1$ if $T_i$ is not censored ($T_i\le C_i$) and $\Delta_i=0$ otherwise, and $X_i(t)=\big(X_i^{(1)}(t),\cdots,X_i^{(p)}(t)\big)$ are covariates observed\footnote{If $X_i(t)$ is sampled at discrete (but possibly irregular) time points, \softa interpolates its trajectory with piecewise constant paths.} from enrollment time ($t=0$) up to $\tilde T_i$.

As a concrete example, suppose that each unit is a patient monitored in an ICU, and the event of interest is mortality in the ICU. A patient is censored if he/she is discharged alive, and $\tilde T$ is the minimum of the time-to-mortality and the time-to-discharge, as measured from the time of admission. The components of $X_i(t)$ could be blood pressure, heart rate, temperature, laboratory test results, and nursing assessments \citep{ma2019}. The hazard $\lambda(t,X_i(t))$ in this case is informally the probability of dying in the ICU in the near future based on patient $i$'s status $X_i(t)$ at time $t$.\footnote{It is also possible to use recent status history $\{X(s)\}_{s\in[t-\tau,t]}$ instead of just current status $X(t)$. See for example \citet{adelson2017} and \citet{ma2019} on how to transform status history into a time-dependent covariate.}

Writing $F(t,x)=\log\lambda(t,x)$ as the log-hazard function, \softa estimates $F(t,x)$ by using functional gradient boosting to minimize the negative log-likelihood functional (the likelihood risk)
\begin{align}
\notag & R_n(F)\\
=&\frac{1}{n}\sum_{i=1}^{n}\left\{ \int_{0}^{\tilde T_i}e^{F(t,X_{i}(t))}dt-\Delta_i F(\tilde T_i,X_i(\tilde T_i))\right\} \label{eq:loglik}
\end{align}
for $F$ in the span of regression trees. Interestingly the functional gradient of $R_n(F)$ cannot be directly computed from \eqref{eq:loglik} because of the way it depends on the functional datapoints $\{X_i(t)_{t\le \tilde T_i} \}_{i=1}^n$. \citet{lee2017boosted} resolves this issue by proving a smooth convex representation for $R_n(F)$, and deriving its functional gradient for arbitrary learner classes. The computational algorithm for \softa described in the next section specializes these results to regression tree learners in order to propose a novel approach for growing the trees.

\section{The \softa Algorithm}
Algorithm \ref{Boosting_overall} describes the \softa algorithm for estimating $\lambda(t,x)$. As an overview, \softa creates a sequence $g_0(t,x),g_1(t,x),\cdots$ of regression tree functions of time $t$ and covariates $x$ in an iterative manner, and uses them to form an ensemble estimate for the log-hazard function 
$$
F_M(t,x) = F_0 - \nu\sum_{m=0}^{M-1} g_{m}(t,x),
$$
from which the hazard estimator can be obtained as $\hat\lambda(t,x) = e^{F_M(t,x)}$. Here, $M$ is the number of boosting iterations, and the default learning rate $\nu=0.1$ is commonly used in boosting applications. The initial guess for the log-hazard, $F_0=\log\frac{\sum_{i=1}^{n}\Delta_{i}}{\sum_{i=1}^{n}\tilde T_{i}}$, is the best constant that minimizes \eqref{eq:loglik}, where the ratio is the total number of observed events divided by the total amount of time at-risk. The number of boosting iterations $M$ as well as the maximum number of splits $L$ in each tree are hyperparameters that are chosen via $K$-fold cross-validation.

Given $F_m(t,x)$, the next iterate $F_{m+1}(t,x)$ is computed by seeking a regression tree $g_m(t,x)$ to add to $F_m(t,x)$, i.e. $F_{m+1}\leftarrow F_{m}-\nu g_{m}$. In gradient boosting, $g_m(t,x)$ needs to be aligned to the gradient function of $R_n(F)$ at $F_m$. This leads to a reduction in $R_n(F)$ from one iteration to the next. The following subsection describes how \softa constructs $g_m(t,x)$ to reduce $R_n(F)$ in a more direct manner than the prototype implementation in \citet{lee2017boosted}.

\begin{algorithm}
\caption{\softa}
\label{Boosting_overall}
\hspace*{\algorithmicindent}
\begin{algorithmic}[1]
\STATE {\textbf{Input:} $n$ functional data samples $\{X_{i}(t)_{t\le\tilde T_{i}},\tilde T_{i},\Delta_{i}:i=1,\ldots,n\}$, \#iterations $M$, maximum tree splits $L$, and default stepsize $\nu=0.1$.}
\STATE {Initialize $F_{0}=\log(\sum_{i=1}^{n}\Delta_{i}/\sum_{i=1}^{n} \tilde T_{i})$.}
\FOR{$m=0$ \TO $M-1$}
\STATE{Initialize $g_{m,0}=0$.}
\FOR{$l=1$ \TO $L$}
\STATE {Identify the tree split ($B_{m,j},A_1,A_2$) that minimizes the score $d$ in \eqref{eq:splitscore}.}
\IF{$d<0$}
\STATE {Calculate ($\gamma_1,\gamma_2$) from \eqref{eq:best_coef}}.
\STATE {Update tree: $g_{m,l}(t,x)=g_{m,l-1}(t,x)-c_{m,j}I_{B_{m,j}}(t,x)+\gamma_{1}I_{A_{1}}(t,x)+\gamma_{2}I_{A_{2}}(t,x)$}.
\ELSE 
\STATE \textbf{break}
\ENDIF
\ENDFOR
\STATE{Compute $F_{m+1}\leftarrow F_{m}-\nu g_{m}$}.
\ENDFOR
\STATE {\textbf{Output:} $\hat\lambda (t,x) = e^{F_M(t,x)}$.}
\end{algorithmic}
\end{algorithm}

\subsection{Constructing The Tree $g_m(t,x)$}
At the $m$-th iteration we seek a regression tree $g_m(t,x)$ that is aligned to the gradient. Since $R_n(F)$ is convex \citep{lee2017boosted}, the alignment property holds for any $g(t,x)$ that leads to a decrease in $R_n(F_m)$ when $F_m$ is moved in the direction of $-g(t,x)$. Moreover, the larger the decrease is, the greater the alignment. This key insight behind \softa leads to a greedy approach for growing $g_m(t,x)$ that is similar to the spirit of \xgb \citep{xgboost}: Starting with a tree with a root node, we choose the split that maximally reduces the likelihood risk, and repeat the process iteratively on successive leaf nodes until the tree has been split (up to) $L$ times. To explain further, let the tree with just a root node be the constant function $g_{m,0}(t,x)=0$, and let
$$g_{m,l}(t,x)=\sum_{j=1}^{l+1}c_{m,j}I_{B_{m,j}}(t,x)$$
be the intermediate tree after $l$ splits. Here, the splits can be on the covariates or on time. The indicator function $I_{B}(t,x)$ represents whether $(t,x)=(t,x^{(1)},\cdots,x^{(p)})$ belongs in the time-covariate hypercube of the form
\begin{equation}
B=\left\{ \begin{array}{ccc}
(t,x) & : & \begin{array}{c}
\underline{t}^{B}<t\leq\overline{t}^{B}\\
\underline{x}^{(1,B)}<x^{(1)}\leq\overline{x}^{(1,B)}\\
\vdots\\
\underline{x}^{(p,B)}<x^{(p)}\leq\overline{x}^{(p,B)}
\end{array}\end{array}\right\} .\label{eq:regioncubes}
\end{equation}
$B_{m,1},\cdots,B_{m,l+1}$ are disjoint regions from the first $l$ splits,
each representing one of the leaf nodes in the intermediate tree. $c_{m,1},\cdots,c_{m,l+1}$ are the values of the tree function in those leaf nodes. To obtain $g_{m,l+1}(t,x)$, we split one of the $B_{m,j}$ regions into two subregions $A_{1}$ and $A_{2}$ of the same form as (\ref{eq:regioncubes}) to get
$$
\begin{aligned}
	g_{m,l+1}(t,x)&=g_{m,l}(t,x)-c_{m,j}I_{B_{m,j}}(t,x)\\
	&\eqindent+\gamma_{1}I_{A_{1}}(t,x)+\gamma_{2}I_{A_{2}}(t,x).
\end{aligned}
$$
The region $B_{m,j}$ to split, the variable or time axis to split on and the location of the split, and also the values of $(\gamma_{1},\gamma_{2})$ are all chosen to minimize $R_n(F_m - g_{m,l+1})$. 

\textit{Departure from the prototype implementation in \citet{lee2017boosted}.} The splits in Section 4 of \citet{lee2017boosted} are chosen to maximize the alignment between the tree and the gradient function of $R_n(F)$. As we know from traditional gradient boosting, subtracting a tree that is closely aligned to the gradient should reduce $R_n(F)$. By contrast, \softa splits are chosen to directly minimize $R_n(F)$, something that our novel algorithm makes possible, and this leads to more targeted risk reduction. Thus the difference between the two implementations is analogous to the difference between \xgb and the traditional boosting approach as they pertain to regression/classification problems. Given the performance gain \xgb enjoys over the traditional approach, we expect \softa to also perform better than the prototype in \citet{lee2017boosted}. We empirically validate this in Section \ref{sec:numerical}.

Whereas \xgb minimizes a second order Taylor approximation to a risk function in order to speed up computations, an innovation of \softa is in discovering how to directly minimize the exact form of $R_n(F)$ in an efficient way: Since the tree values $\gamma_1,\gamma_2$ only apply to the subregions $A_1, A_2$, we can write $R(F_{m}-g_{m,l+1})$ as
\[
\begin{aligned}
& \frac{1}{n} \sum_{i=1}^n \sum_{k=1}^{2}\Biggl\{ \int_0^{\tilde T_i}e^{F_{m}(t,X_{i}(t)) - \gamma_k }I_{A_{k}}(t,X_{i}(t))dt \\
& \quad - \Delta_{i} \left[F_{m}(\tilde T_i, X_i(\tilde T_i)) - \gamma_k\right] I_{A_k}(\tilde T_i, X_i(\tilde T_i)) \Biggr\}+C \\
= & \frac{1}{n} \sum_{i=1}^n \sum_{k=1}^{2}\Biggl\{ e^{-\gamma_k} \cdot \int_0^{\tilde T_i}e^{F_{m}(t,X_{i}(t))}I_{A_{k}}(t,X_{i}(t))dt \\
& \qquad + \gamma_k \cdot \Delta_{i} I_{A_k}(\tilde T_i, X_i(\tilde T_i)) \Biggr\} + C' \\
\end{aligned}
\]
where $C,C'$ are quantities that do not depend on $\gamma_1$ or $\gamma_2$. Rewriting the above yields
\begin{equation}
R_n(F_{m}-g_{m,l+1}) =\sum_{k=1}^{2} [e^{-\gamma_{k}}U_k+\gamma_{k}V_k]+C',
\label{eq:lik_risk_expand}
\end{equation}
\begin{equation}
\begin{aligned}
U_k&=\frac{1}{n}\sum_{i=1}^{n}\int_0^{\tilde T_i}e^{F_{m}(t,X_{i}(t))}I_{A_{k}}(t,X_{i}(t))dt,\\
V_k&=\frac{1}{n} \sum_{i=1}^n \Delta_i I[ \{\tilde T_i, X_i(\tilde T_i)\} \in A_k ].
\end{aligned}
\end{equation}
Note that $V_k$ is the (scaled) number of events observed in $A_k$. If $U_k,V_{k}>0$ then the minimizing value in $A_k$ is
\begin{equation}
\gamma_{k}=\log(U_{k}/V_{k}).\label{eq:best_coef}
\end{equation}
Putting this into the expression for $R(F_{m}-g_{m,l+1})$
above yields the optimized value
\begin{equation}
R(F_{m}-g_{m,l+1})=\sum_{k=1}^{2}V_{k}\left(1+\log\frac{U_{k}}{V_{k}}\right)+C'.\label{eq:best_R}
\end{equation}
By reasoning inductively, the decrease in the likelihood risk due
to the new split is
\begin{equation}
\begin{aligned}
d&=R(F_{m}-g_{m,l+1})-R(F_{m}-g_{m,l})\\
&=V_{1}\left(1+\log\frac{U_{1}}{V_{1}}\right)+V_{2}\left(1+\log\frac{U_{2}}{V_{2}}\right)\\
&\eqindent-(V_{1}+V_{2})\left(1+\log\frac{U_{1}+U_{2}}{V_{1}+V_{2}}\right),
\end{aligned}
\label{eq:splitscore}
\end{equation}
which can be viewed as a score for determining the best split: Of all possible splits (defined by $B_{m,j}$, $A_{1}$, and $A_{2}$) where $U_k,V_k>0$ in both subregions, the best one is that which minimizes \eqref{eq:splitscore}. Since at each iteration only two new leaf nodes are created, it is only necessary to determine the best split for each of the two new regions at the next iteration: $d$ remains unchanged for the other leaf nodes from previous iterations. If $d\ge 0$ for all leaf nodes then we stop splitting because doing so will not reduce the likelihood risk further.  

One difference between growing a tree for the traditional non-functional data setting versus the functional data setting (our setting) is that for the latter, when choosing a variable to split on, time $t$ is also treated as a candidate variable just like $x^{(1)},\cdots,x^{(p)}$.

%
%

\subsubsection{Candidate Split Points for Variables}
Given a set of candidate split points for a particular variable, the best split point is that which minimizes \eqref{eq:splitscore}. If $x^{(j)}$ is continuous, \softa proposes split candidates based on the percentiles of the observed data for $x^{(j)}$. The default setting places a candidate split at every decile.
 
If $x^{(j)}$ is categorical, \softa employs a one-hot encoding heuristic: Set $A_{1}$ equal to the intersection of $B_{m,j}$ with the region
where $x^{(j)}$ equals a particular categorical label. Hence $A_{2}$ is the intersection of $B_{m,j}$ with the region where $x^{(j)}$ is any other label. The algorithm would then choose the category label for $A_{1}$ that minimizes the score \eqref{eq:splitscore}. The rationale for this is that if $U_{k}$ and $V_{k}$ can be varied continuously, then \eqref{eq:splitscore} tends to $-\infty$ as the
ratio $U_{1}/V_{1}$ tends to $\infty$. Hence a heuristic is to find a subset of categorical labels to intersect with $B_{m,j}$ so that $U_{1}/V_{1}$ is maximized. This always has a solution in the form of a singleton set.
 
By contrast, the prototype implementation in \citet{lee2017boosted} chooses splits to minimize the mean squared error between the tree and gradient. As such, \softb's splitting rules for both continuous and categorical variables are more directly targeted at reducing $R_n(F)$.

\subsubsection{Imputing Covariate Trajectories}
When the paths of $X(t)$ are continuous, no subregion $A_k$ can contain an observed event unless it is traversed by at least one covariate trajectory, i.e. $V_k>0 \Rightarrow U_k>0$. This ensures that, as long as we consider only candidate splits where the number of observed events is positive in both subregions, $\gamma_k=\log(U_k/V_k)$ will be well defined.

\begin{figure}[t!]
\centering
\begin{subfigure}[t]{0.7\columnwidth}
   \includegraphics[width=1\linewidth]{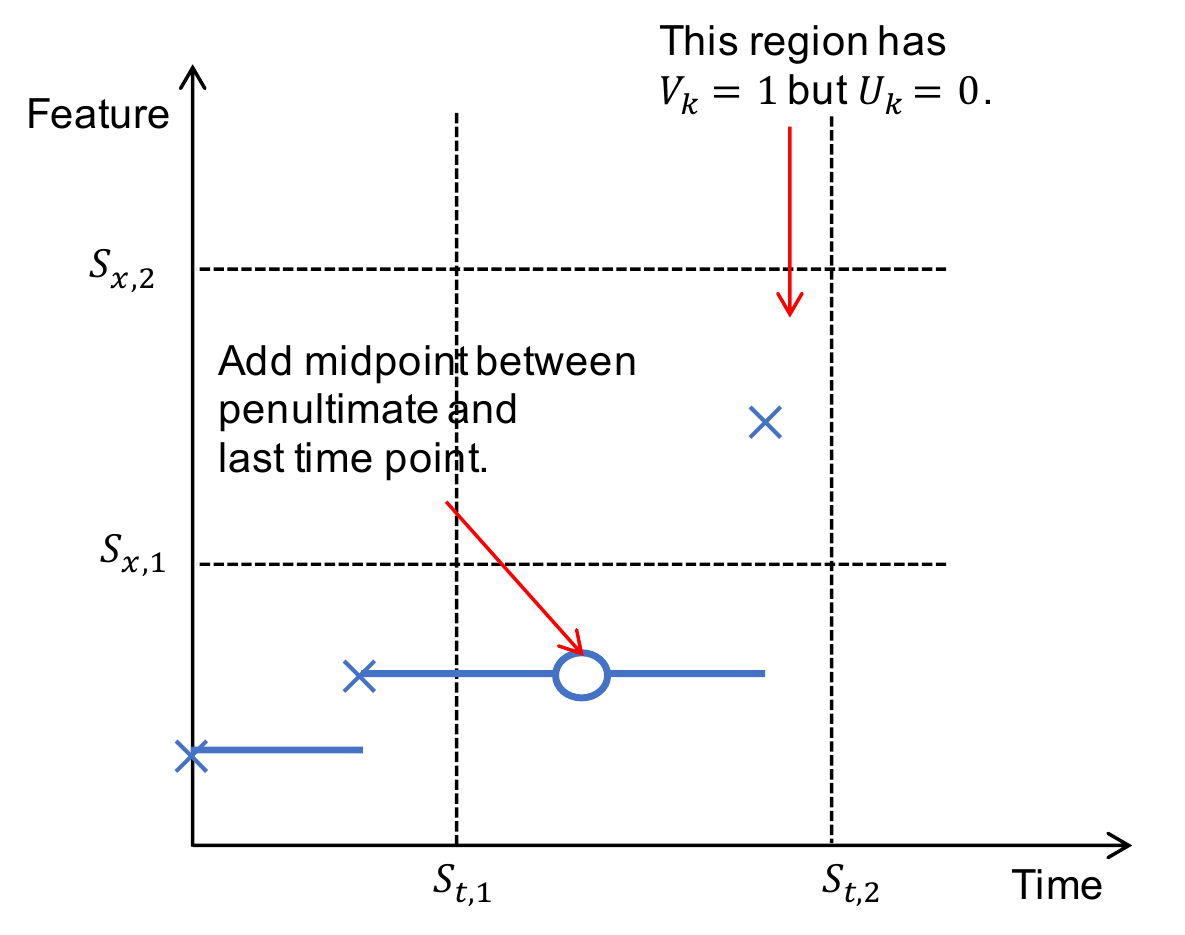}
   \caption{}
   \label{fig:uk_original} 
\end{subfigure}

\begin{subfigure}[t]{0.7\columnwidth}
   \includegraphics[width=1\linewidth]{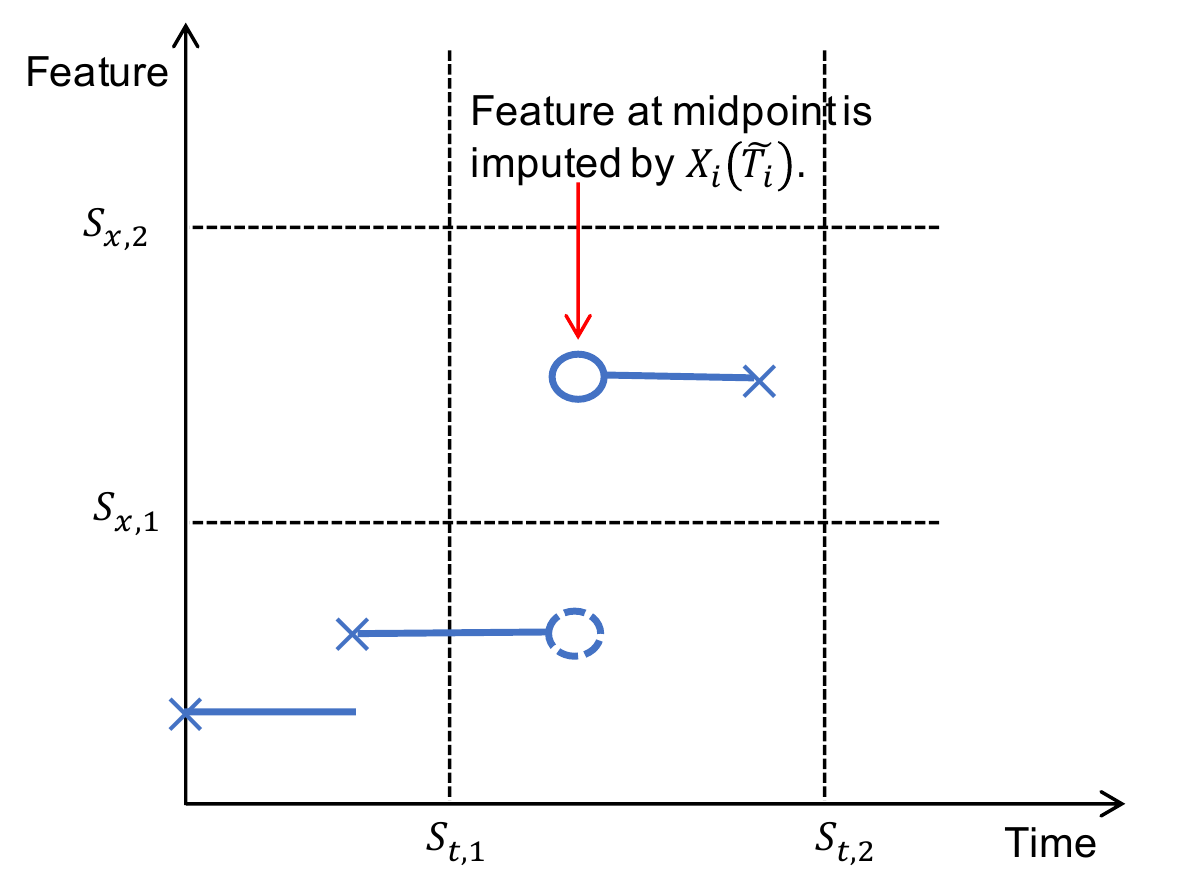}
  \caption{}
   \label{fig:uk_processed}
\end{subfigure}
\caption{(a) Discontinuous Covariate Trajectory Resulting In $V_k=1$ and $U_k=0$ For One Of The Time-Covariate Regions. (b) Imputed Trajectory Mitigates This Problem.}
\vskip -0.2in  
\end{figure}

However, when the paths of $X(t)$ are discontinuous, it is possible for $V_k>0$ while $U_k=0$. In Figure \ref{fig:uk_original}, the trajectory jumps before time $S_{t,1}$ (the first split point on the time axis), and jumps again just before the second split $S_{t,2}$, whereupon the observation experiences the event at the rightmost \textcolor{blue}{$\times$}. Here, the region bounded by ($S_{t,1},S_{t,2}$) on the time axis and by ($S_{x,1},S_{x,2}$) on the covariate axis, has $V_k=1/n$ but $U_k=0$. To mitigate this, we impute a jump at the midpoint between the penultimate and the last time points (\textcolor{blue}{O} in Figure \ref{fig:uk_processed}), and assign the values of the covariates from the last timepoint, $X_i(\tilde T_i)$, to the imputed one.

\subsection{Defining Variable Importance}
A variable importance measure can be constructed for the \softa estimator: Define the importance of the $k$-th variable (the zero-th one being time $t$) as
\begin{equation}
\mathcal{I}_{k}=\sum_{m=0}^{M-1}\mathcal{I}_{k}(g_{m}),\label{eq:vimp}
\end{equation}
where for tree $g_{m}$ with $L$ internal nodes,
\[
\mathcal{I}_{k}(g_{m})=-\sum_{\ell=1}^{L}d_{\ell}I(v(\ell)=k)\geq0.
\]
Here, $d_\ell$ is the split score \eqref{eq:splitscore} at iteration $\ell$ and $v(\ell)$ is the variable used for the partition. Hence the second sum represents
the total reduction in likelihood risk due to splits on the $k$-th variable in the $m$-th tree, and $\mathcal{I}_k$ is the total reduction across the $M$ trees. To convert $\mathcal{I}_{k}$ into a measure of relative
importance between 0 and 1, it is scaled by $\max_{k}\mathcal{I}_{k}$, where a larger value confers higher importance. Since the prototype implementation in \citet{lee2017boosted} fits trees to the gradients of $R_n(F)$ instead, it defines variable importance as the reduction in mean squared error between the trees and the gradients. Therefore, \softb's variable importance is a more direct measure of a variable's contribution to reducing the likelihood risk.


\section{NUMERICAL STUDY}\label{sec:numerical}
We now compare the performance of \softa to those of several existing methods. For this, we use simulated datasets for which the true hazard function is known, thus allowing us to compute how well the methods do in recovering the truth. We then apply \softa to the Framingham Heart Study dataset to uncover a novel clinical finding concerning the relationship between blood pressure and CVD.

\begin{table*}[t]
\vskip 0.15in
    \caption{$\text{err}_{L^2}$ with $95\%$ confidence intervals for the simulated datasets (smaller values are better). Numbers rounded to two significant figures. \softb's hyper-parameters are tuned to the training data, whereas those for the other methods are tuned directly to the test data, Note that this puts \softa at an disadvantage. Furthermore, \flexsurv includes the log-normal distribution as one of its parametric options, so it is correctly specified for $\lambda_3$.} 	
    \label{tab:results_synthetic}
    \vskip 0.15in
    \begin{center}
    \begin{small}
    \begin{tabular}{cc|cccc}
    \hline
    \multirow{2}{*}{Hazard} & \multirow{2}{*}{\#Irrelevant covariates} &\multicolumn{4}{c}{Estimator}  \\
    & & \softa & \kernel &	\flexsurv  &	\blackboost	\\\hline
    \multirow{3}{*}{$\lambda_1$}& 0  & 0.17 (0.17, 0.17) & \textbf{0.14 (0.14, 0.15)}    & 0.53 (0.52, 0.54) & 0.58 (0.57, 0.59) \\
                                & 20 & \textbf{0.20 (0.20, 0.20)} & 3.4 (3.0, 3.9)    & 0.54 (0.53, 0.54) & 0.58 (0.57, 0.59) \\
                                & 40 & \textbf{0.21 (0.20, 0.21)} & 43 (5.7, 80)  & 0.54 (0.54, 0.55) & 0.58 (0.57, 0.59) \\ \hline
    \multirow{3}{*}{$\lambda_2$}& 0  & 0.23 (0.23, 0.24) & \textbf{0.11 (0.11, 0.12))}    & 1.1 (1.1, 1.1) & 1.4 (1.4, 1.4) \\
                                & 20 & \textbf{0.25 (0.25, 0.26)} &  4.5 (3.9, 5.2)    & 1.1 (1.1, 1.1) & 1.4 (1.4, 1.4) \\
                                & 40 & \textbf{0.26 (0.26, 0.27)} & 29 (11, 46) & 1.1 (1.1, 1.1) & 1.4 (1.4, 1.4) \\ \hline
    \multirow{3}{*}{$\lambda_3$}& 0  & 0.038 (0.037, 0.040) & 0.046 (0.044, 0.049)    & \textbf{0.0040 (0.0039, 0.0041)} & 0.10 (0.10, 0.11) \\
                                & 20 & 0.047 (0.046, 0.049) & 1.8 (1.1, 2.5)  & \textbf{0.020 (0.019, 0.020)} & 0.10 (0.10, 0.11) \\
                                & 40 & 0.050 (0.048, 0.051) & 7.6 (5.3, 9.7)    & \textbf{0.030 (0.029, 0.031)} & 0.10 (0.10, 0.11) \\ \hline
    \multirow{3}{*}{$\lambda_4$}& 0  & 0.049 (0.048, 0.050) & \textbf{0.045 (0.044, 0.046)}    & 0.20 (0.19, 0.20) & 0.20 (0.19, 0.20) \\
                                & 20 & \textbf{0.060 (0.059, 0.062)} & 3.9 (0.66, 7.1)    & 0.20 (0.19, 0.20) & 0.20 (0.19, 0.20) \\
                                & 40 & \textbf{0.069 (0.067, 0.070)} &   5.5 (4.3, 6.7)    & 0.20 (0.20, 0.21) & 0.20 (0.19, 0.20) \\ \hline
    \end{tabular}
    \end{small}
    \end{center}
\vskip -0.1in
\end{table*}

\begin{table*}[t]
\vskip 0.15in
    \caption{\softa vs. the prototype implementation in \citet{lee2017boosted}: Performance and speed comparisons. On average, \softa is 16\% faster than the prototype and also achieves a 5.1\% reduction in $L^2$-error. Numbers rounded to two significant figures.}
    \label{tab:boxhed_vs_lee}
    \vskip 0.15in
    \begin{center}
    \begin{small}
    \begin{tabular}{cc|cc}
    \hline
    \multirow{2}{*}{Hazard} & \multirow{2}{*}{\#Irrelevant covariates} &\multicolumn{2}{c}{Percentage reduction achieved by \softa}  \\
    & & Error $\text{err}_{L^2}$ & Computation time \\ \hline
    \multirow{3}{*}{$\lambda_1$}& 0  & -0.35\% & -0.52\% \\
                                & 20 & 3.7\% & 17\% \\
                                & 40 & 3.0\% & 0.72\% \\ \hline
    \multirow{3}{*}{$\lambda_2$}& 0  & 1.9\% & 48\% \\
                                & 20 & 2.5\% & -3.4\% \\
                                & 40 & 3.0\% & 16\% \\ \hline
    \multirow{3}{*}{$\lambda_3$}& 0  & 5.0\% & 0.78\% \\
                                & 20 & 5.2\% & 1.1\% \\
                                & 40 & 5.0\% & 0.97\% \\ \hline
    \multirow{3}{*}{$\lambda_4$}& 0  & 27\%  & 57\% \\
                                & 20 & 1.8\% & 33\% \\
                                & 40 & 3.5\% & 15\% \\ \hline
    \multicolumn{2}{c}{\textbf{Average \softa improvement} (95\% C.I.)}  &\textbf{5.1\%} (1.1\%, 9.1\%) & \textbf{16\%} (4.0\%, 27\%) \\ \hline
    \end{tabular}
    \end{small}
    \end{center}
\vskip -0.1in
\end{table*} 

\subsection{Performance Metrics}
We evaluate the performance of the estimators on test data using two metrics: $L^2$-error and time-dependent AUC ($AUC_t$). Details on the training/test data splits can be found in Section \ref{data_descrip}.


The $L^2$-error is calculated on test datasets of $N$ randomly sampled data points, and is defined as
$$
\text{err}_{L^2} = \left\{ \frac{1}{N}\sum_{i=1}^{N}(\hat{\lambda}_i-\lambda_i)^2 \right\}^{1/2},
$$
where $\hat{\lambda}_i$ and $\lambda_i$ are the predicted and true hazard values for the $i$-th test data point. Note that $\text{err}_{L^2}$ is always non-negative, and a smaller value indicates higher accuracy. In particular, $\text{err}_{L^2}=0$ if and only if the predictions are perfect, i.e. all the predicted values are exactly the same as the true hazard values. 

$AUC_t$ is defined as follows \citep{blanche2019c}:
$$AUC_t = P \left(\hat{S}_i(t)<\hat{S}_j(t) \vert \Delta_i=1,  T_i <t< T_j \right),$$
where $\hat{S}_i(t)$ is the conditional survival probability given the covariate trajectory $\{X_i(s):0\leq s\leq t\}$.  Note that $AUC_t$ lies between $0$ and $1$, and a value of $0.5$ corresponds to a random guess. A larger value indicates better performance. However, $AUC_t$ is not as sharp as $\text{err}_{L^2}$ in detecting prediction mistakes. For example, if we overpredict by a factor of 2, i.e. $\hat \lambda_i = 2\lambda_i$, the corresponding $AUC_t$ will still attain the best possible value of $1$, whereas $\text{err}_{L^2}$ will strictly be larger than zero.




\subsection{Baseline Comparisons}
We compare \softa to several existing hazard estimation methods: Kernel smoothing \citep{neilsen}, parametric hazard estimators for time-dependent covariates (\flexsurv in R), and boosted parametric estimators for time-static covariates (\blackboost in R). The Cox proportional hazards model is excluded because it is only able to estimate the cumulative hazard but not the hazard itself. The hyper-parameters\footnote{Candidates $L\in\{1, 2, 3, 4\}$ and $M\in\{100, 150, \cdots, 300\}$.} for \softa are tuned using five-fold cross-validation on the training data. For kernel smoothing we utilize the kernel function $K(u) = \frac{3003}{2048}(1-x^2)^6I(-1<x<1)$ from \citet{perez}. The hyper-parameters\footnote{Bandwidth for {\ttfamily kernel}, choice of parametric family for \flexsurv and {\ttfamily blackboost}, and the number of trees in {\ttfamily blackboost}.} for the baseline estimators are tuned directly to the test data (see Appendix for details). Note that this puts \softa at a significant disadvantage.

To assess the performance gain from using our novel tree splitting rule, we also compare \softa to the prototype implementation in \citet{lee2017boosted}. Since no public implementation exists for the latter, we re-implemented a version of it to use for comparison.

\subsection{Data Description} \label{data_descrip}
\textit{Simulated datasets.}
We consider four datasets simulated from the following hazard functions used in \citet{perez}, with $x_t$ being a piecewise-constant function with values drawn from $U(0,1]$: 
\begin{equation*}
\begin{aligned}
\lambda_{1}(t,x_t)&=B(t,2,2)\times B(x_t,2,2),t\in (0, 1],\\
\lambda_{2}(t,x_t)&=B(t,4,4)\times B(x_t,4,4),t\in (0, 1],\\
\lambda_{3}(t,x_t)&=\frac{1}{t}\frac{\phi(\log t-x_t)}{\Phi(x_t-\log t)},t\in(0,5],\\
\lambda_{4}(t,x_t)&=\frac{3}{2}t^{\frac{1}{2}}\exp\left(-\frac{1}{2}\cos(2\pi x_t)-\frac{3}{2}\right), t\in(0,5],\\
\end{aligned}
\end{equation*}
where $B(\cdot, a, a)$ is the PDF of the Beta distribution with shape and scale parameters equal to $a$, and $\phi(\cdot)$ and $\Phi(\cdot)$ denote the PDF and CDF of $N(0,1)$. In other words, $\lambda_1$ and $\lambda_2$ take the form of Beta PDFs, and $\lambda_3$ is the hazard of the log-normal distribution. As will be explained in Section \ref{sim_results}, the first two cases naturally favour \kernel while the third one favours \flexsurvb.

To investigate the robustness of \softa to noise in a high dimensional setting, we also add up to 40 irrelevant covariates to each hazard function. The trajectories of the covariates are simulated as piecewise-constant paths with values drawn from $N(0,1)$.

For the training set we draw 5,000 sample trajectories, and we also draw 5,000 for the test set.

\textit{Framingham Heart Study dataset.}
We pool together longitudinal records from two prospective cohorts: The Framingham Heart Study original cohort (FHS) and the Framingham Heart Study Offspring Cohort (FHS-OS) \citep{dawber1951epidemiological}. The event of interest is the time to first onset of cardiovascular disease (CVD). We apply \softa to the data to identify important risk factors and to also uncover novel interaction effects. This might help us better understand the science behind CVD progression, and also to identify high-risk individuals for early intervention.

The pooled data consists of 9,697 participants and 73,340 physical exam records. Eight risk factors were consistently collected across all exams, and these are used in common medical models: Age, gender, systolic blood pressure (SBP), diastolic blood pressure (DBP), smoking status, diabetes, total cholesterol (TC), and body mass index (BMI). The 9,697 study participants are randomly split into 7,000/2,697 for training/testing.  Additional details on cohort selection can be found in the Appendix.

%

\subsection{Performance On Simulated Datasets} \label{sim_results}
Table \ref{tab:results_synthetic} presents the $L^2$-errors for the hazard estimators when applied to the simulated datasets. Several observations are in order:
\begin{itemize}
\item \softa always outperforms \kernel when irrelevant covariates are present. The methods are comparable when no irrelevant covariates are present, with \kernel having the edge in the first two cases $\lambda_1$ and $\lambda_2$. This is because the kernel function $K(u)$ is a location- and scale-transformed Beta PDF, which is also the functional form for $\lambda_1$ and $\lambda_2$. It is therefore unsurprising that \kernel is able to approximate $\lambda_1$ and $\lambda_2$ better than regression trees. The minuscule edge that \kernel enjoys for $\lambda_4$ is likely due to the fact that is was tuned directly to the test data.
\item For $\lambda_3$, \flexsurv performs the best, followed closely by \softb. The reason for \flexsurvb's outperformance is due to the fact that it includes the log-normal distribution as one of its parametric options, so it is correctly specified for $\lambda_3$.
\item Neither \softa nor \blackboost are affected much by irrelevant covariates, while \kernelb's performance drops dramatically when irrelevant covariates are added. These findings are in line with the fact that kernel smoothing suffers from the curse of dimensionality, while boosted trees automatically perform variable selection.
\end{itemize}

\begin{figure}[th]
\centering
\begin{subfigure}[t]{0.9\columnwidth}
   \includegraphics[width=1\linewidth]{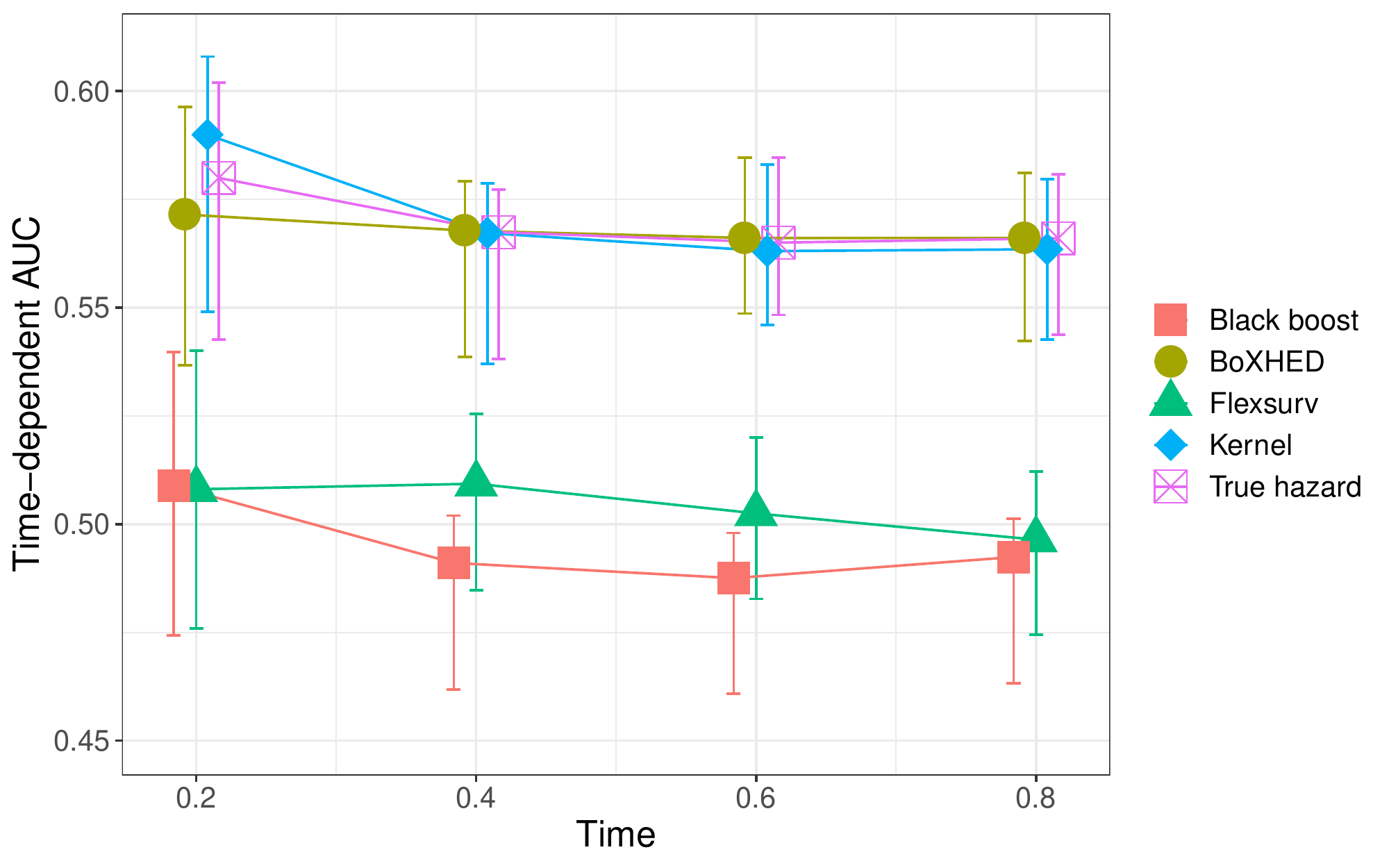}
   \caption{}
   \label{fig:plot_AUC_irr0} 
\end{subfigure}
\vskip 0.1in  

\begin{subfigure}[t]{0.9\columnwidth}
   \includegraphics[width=1\linewidth]{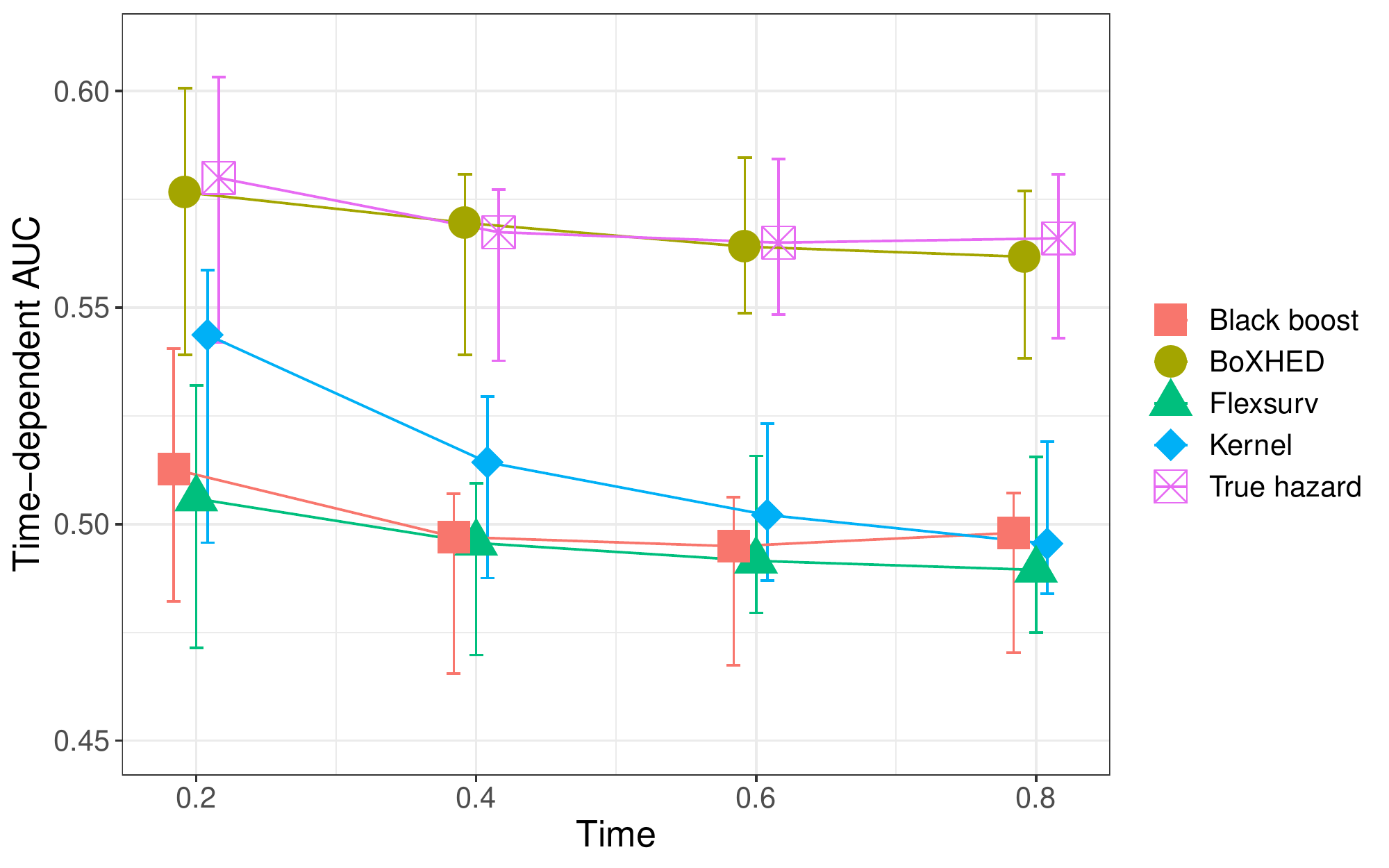}
  \caption{}
   \label{fig:plot_AUC_irr20}
\end{subfigure}
\caption{$AUC_t$ versus time $t$ for the estimators when applied to data simulated from $\lambda_1$. Larger $AUC_t$ values are better. (a) No irrelevant covariates; (b) 20 irrelevant covariates.}
\label{fig:plot_AUC}
\vskip -0.2in  
\end{figure}

Figure \ref{fig:plot_AUC_irr0} presents the $AUC_t$ results for the estimators when applied to data simulated from $\lambda_1$ (no irelevant covariates). The performances of \softa and \kernel are statistically indistinguishable from that of the true hazard function, while \blackboost and \flexsurv perform no better than random guessing. When we increase the number of irrelevant covariates to just 20 (Figure \ref{fig:plot_AUC_irr20}), even \kernel becomes statistically indistinguishable from random guessing. In terms of $AUC_t$, \softa remains as the only method that is able to perform as well as the true hazard. Similar results for $\lambda_2$, $\lambda_3$, and $\lambda_4$ are shown in the Appendix.

Table \ref{tab:boxhed_vs_lee} compares the performance of \softa to that of the prototype implementation in \citet{lee2017boosted}. We see that \softa outperforms the prototype in both speed and accuracy in $10$ out of the $12$ scenarios, and both approaches perform essentially the same on an eleventh one. On average, \softa achieved a $5.1\%$ lower $L^2$-error ($95\%$ C.I. $1.1\% - 9.1\%$) and takes $16\%$ less time to compute ($95\%$ C.I. $4.0\%-27\%$), excluding the shared data pre-processing time.

\subsection{Analyzing The Framingham Heart Study}
We now use \softa to analyze the Framingham Heart Study dataset in order to identify important risk factors and interactions amongst them. Recall that the event of interest is the time to first onset of CVD.

\subsubsection{Variable Importances}
Table \ref{tab:feature_importance} shows the relative variable importances \eqref{eq:vimp} identified by \softa (scaled from $0$ to $100$). The top risk factors (age, SBP, smoking status, gender) match those identified by existing CVD prediction algorithms widely used by clinicians (e.g. ACA/AHA risk score \citep{arnett20192019}, Framingham risk score \citep{lloyd2004framingham}). 

\subsubsection{Interactions Among Risk Factors}
Since cross-validation selected trees with two splits,
this suggests the presence of two-way interactions among the risk factors and time. We investigate the impact of blood pressure (BP) on CVD risk (defined as the hazard of CVD onset), and show that \softa is able to uncover clinically novel interaction effects. 

To systematically examine the relationship between BP and CVD risk, we calculate the hazard estimated by \softa for $576$ sub-cohorts. A sub-cohort is defined by combinations of $7$ variables: Age (50, 60, and 70 years old), gender (women, men), TC (5, 5.7, 6.5 $mmol/L$), current smoker (yes, no), diabetes (yes, no), BMI (20, 25, 30, and 35 $kg/m^2$), and cohort indicator (FHS, FHS-OS). For each sub-cohort, the estimated hazard is computed for different values of BP: $\text{SBP}\in [90, 170]$ mmHg and $\text{DBP}\in [60, 110]$ mmHg. The estimated hazard surfaces are scaled to $[0, 1]$ and are aggregated into four clusters using $K$-means clustering (see Appendix for details).

\begin{table}[t]
    \caption{Relative importances (out of 100) and $95\%$ confidence intervals (CI) for the risk factors in the Framingham Heart Study. Confidence intervals obtained from 1,000 bootstrapped datasets.}
    \label{tab:feature_importance}
    \vskip 0.15in
    \begin{center}
    \begin{small}
    \begin{tabular}{lc}
    \hline
	Age &100\\
	SBP&15 (11,21)\\
	Smoking&9.5 (6.3,12)\\
	Gender&8.3 (5.8,11)\\
	Diabetes&6.2 (4.1,8.8)\\
	DBP&3.5 (2.5,8.4)\\
	\tabincell{l}{Total \\Cholesterol}&2.8 (2.1,6.0)\\
	BMI&1.4 (1.2,3.5)\\\hline
    \end{tabular}
\end{small}
    \end{center}
    \vskip -0.1in
\end{table}

\begin{figure}[t!] 
\vskip 0.2in
   \centering
   \includegraphics[width = 0.4\textwidth]{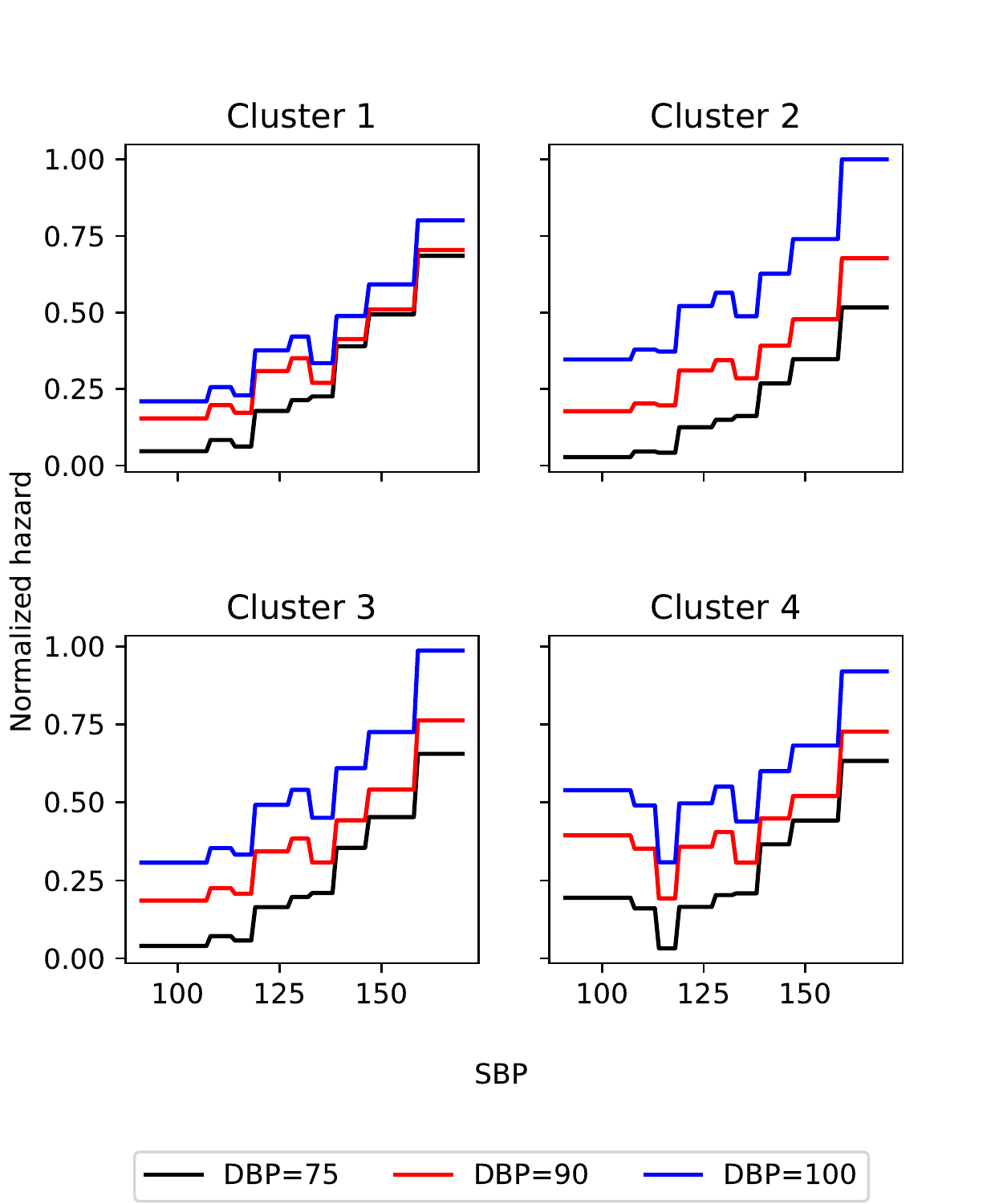} 
   \caption{Plots of hazard against SBP for centroids of clusters.}
   \label{fig:normalized_SBP}
\vskip -0.2in
\end{figure}

Figure \ref{fig:normalized_SBP} exhibits the relationship between SBP and the scaled hazard for the centroid of each cluster, while also conditioning on DBP at three different values (75, 90, and 100 mmHg). Interestingly, for Clusters 1 to 3, CVD risk is increasing in SBP for a given level of DBP, while for Cluster 4 the relationship is U-shaped. In order to better understand the underlying interaction effect, in Table \ref{tab:risk_factor_normalized_hazard} we summarize the characteristics of the four clusters. Age, \%smokers, and TC levels are similar across all clusters, while \%diabetes are similar in Clusters 3 and 4. Gender and BMI are the only covariates that differentiate Cluster 4 from the others, and they indicate an all-male cluster with high BMI values. Therefore, we hypothesize that SBP$\times$BMI and/or SBP$\times$Gender might be the interaction effects responsible for the observed relationship between SBP and CVD risk.

\begin{table}[t]

    \caption{Distribution of risk factors by clusters.} 
    \label{tab:risk_factor_normalized_hazard}
    \vskip 0.15in
    \begin{center}
    \begin{small}
    \begin{tabular}{ccccc}
    \hline
    \multirow{2}{*}{Risk Factor}&\multicolumn{4}{c}{Clusters}\\
         &\#$1$  & \#$2$  & \#$3$ & \#$4$ \\\hline
        Age &	60&	60&	60&	60\\
        \textbf{Female} ($\%$)&	\textbf{60}&	\textbf{52}&	\textbf{59}&	\textbf{0}\\
        \tabincell{c}{Current \\Smoker ($\%$)}&	50&	48&	51&	50\\
        Diabetes ($\%$)&	100&	 16&	46&	50\\
        TC ($mmol/L$) &	5.73&	5.72&	5.74&	5.73\\
        \textbf{BMI ($kg/m^2$)} &	\textbf{29.0}&	\textbf{23.4}&	\textbf{27.1}&	\textbf{35.0}\\\hline    
    \end{tabular}
	\end{small}
    \end{center}
\vskip -0.1in
\end{table}

We use logistic regression to test these hypotheses using odds ratios (ORs). ORs quantify the likelihood of developing CVD relative to the baseline cohort (defined here as SBP $<115$ mmHG and DBP $<$ 70 mmHG). An OR greater than $1$ indicates that CVD onset is relatively more likely for the cohort of interest, whereas an OR less than $1$ indicates that the cohort is relatively less risky. Observations from all participants are pooled together to evaluate the eight risk factors, $\text{SBP}\times \text{DBP}$, and the candidate interaction effects SBP$\times$BMI and SBP$\times$Gender. SBP is bucketed into quintiles ($<$115, 115-124, 125-139, 140-149, and $\geq$150 mmHg), and DBP is bucketed in the same way ($<$70, 70-79, 80-84, 85-89, $\geq$90 mmHg).

Table \ref{tab:Odds_ratio} shows the ORs for the nine lower deciles of BMIs ($\leq 32 kg/m^2$) as well as for the top decile ($>32 kg/m^2$). A similar stratification for SBP and Gender can be found in the Appendix. Cells containing fewer than 10 events are left blank since we do not have enough data to estimate the OR. Among the bottom nine deciles of BMIs (Table \ref{tab:Odds_ratio_normal}), reading down a given column reveals an increasing relationship between OR and SBP for a given level of DBP. On the other hand, the top BMI decile (Table \ref{tab:Odds_ratio_obese}) demonstrates a U-shaped relationship, with the minimum OR attained when the SBP is between $115$ to $124$ mmHg. This matches the trough seen in the plot for Cluster 4 in Figure \ref{fig:normalized_SBP}. Repeating the analysis for SBP$\times$Gender did not reveal a similar qualitative difference (see Appendix).

This line of inquiry, spurred by the findings of \softb, suggests that the SBP$\times$BMI interaction effect might be responsible for the observed differences in the qualitative relationship between CVD risk and SBP. This potentially resolves an open question in the clinical literature regarding the differential impact of BP on CVD development among different patient cohorts \citep{herrington2017evidence, franklin2013hypertension}. Indeed, the fact that prior work remove BMI as an explanatory variable (primary and interaction) may be precisely why the literature report seemingly contradictory findings on these relationships. While our analysis is not intended to be confirmatory, it provides clinical researchers with a concrete hypothesis on which to base further confirmatory analysis.

\begin{table}[t]

    \caption{ORs with 95\% confidence intervals.}
    \label{tab:Odds_ratio}
    \begin{center}
    \begin{small}
        \begin{subtable}{0.5\textwidth}
        \caption{Bottom nine deciles of BMIs ($\leq 32 kg/m^2$).} 
    \label{tab:Odds_ratio_normal}
    \begin{center}
    \resizebox{0.9\columnwidth}{!}{%
    \begin{tabular}{c|ccccc}
    \hline
         \diagbox[width=1.6cm, height=.8cm]{SBP}{\raisebox{.5ex}{DBP}}& $<$70 & 70-79  & 80-84 & 85-89 & $>$90 \\\hline
        $<$115     &	1.0     &	 0.9$\pm$0.2     &	               &	            & \\
        115-124  &	1.5$\pm$0.3     &	1.3$\pm$0.2     &	  1.2 $\pm$0.3         &	      &   \\
        125-139  &        1.8$\pm$0.3     &	1.5$\pm$0.3      &	1.3$\pm$0.2            &	1.9$\pm$0.3         & 1.8$\pm$0.4  \\
        140-149  &	 2.0$\pm$0.5     &	2.0$\pm$0.4       &	1.7$\pm$0.4            &	2.0$\pm$0.4    &  2.3$\pm$0.6 \\
        $>$150    &	  2.8$\pm$0.6     &	 2.6$\pm$0.6      &	 2.4$\pm$0.4       &	    2.2$\pm$0.6        &  2.7$\pm$0.6 \\\hline

    \end{tabular}}
    \end{center}    
    \vskip -0.1in
    \end{subtable}
    
    \begin{subtable}{0.5\textwidth}
    \caption{Top decile of BMIs ($> 32 kg/m^2$).}
    \label{tab:Odds_ratio_obese}
    \begin{center} 
      \resizebox{0.9\columnwidth}{!}{%
    \begin{tabular}{c|ccccc}
    \hline
         \diagbox[width=1.6cm, height=.8cm]{SBP}{\raisebox{.5ex}{DBP}}& $<$70 & 70-79  & 80-84 & 85-89 & $>$90 \\\hline
        $<$115    &	2.0$\pm$0.7	 &1.8$\pm$0.7    &	          &	  	&\\
        115-124  &	1.4$\pm$0.5 		& 1.2$\pm$0.4	&1.1$\pm$0.5     &	    	&   \\
        125-139  &	2$\pm$0.5		&1.6$\pm$0.6	&1.5$\pm$0.4	&2.1$\pm$0.5	&2$\pm$0.5 \\
        140-149  &	2$\pm$0.6&	2$\pm$0.5	&1.7$\pm$0.5	&2.1$\pm$0.7	&2.4$\pm$0.6   \\
        $>$150     &	  3.3$\pm$1.0&	3$\pm$0.7&	2.8$\pm$0.7&	2.6$\pm$0.6&	3.1$\pm$0.6  \\\hline
                
    \end{tabular}}
    \end{center} 
    \end{subtable}
	\end{small}
    \end{center}
\vskip -0.1in
\end{table}

\section{CONCLUSION}
Survival data with time-dependent covariates is becoming more prominent within the machine learning for healthcare community. As the first public implementation of a boosted nonparametric hazard estimator for time-dependent covariates, \softb's utility should increase as high frequency medical data becomes more prevalent. With an in-depth analysis of a heart study cohort, we demonstrate that i) \softb's built-in variable selection capabilities make it robust to high-dimensional data with many irrelevant covariates; ii) \softa is able to flexibly capture novel interaction effects among risk factors, potentially resolving an open clinical question; and iii) The information conveyed by the changes in risk factors over time may be valuable in prognosticating the onset of CVD. We have made \softa available to extend the machine learning toolbox for survival analysis.

\section*{ACKNOWLEDGEMENTS}
BJM was supported in part by NIH grant 1R21EB028486-01. We are grateful to Huajie Qian for helpful discussions.

\bibliography{BoXHED_ref}
\bibliographystyle{icml2020}

\end{document}